\documentclass[letterpaper]{article} 
\usepackage{aaai25}  
\usepackage{algorithm}
\usepackage{algorithmic}
\usepackage{amsmath}
\usepackage{amssymb}
\usepackage{booktabs}
\usepackage{caption} 
\usepackage{courier}  
\usepackage{graphicx} 
\usepackage{helvet}  
\usepackage[hyphens]{url}  
\usepackage{listings}
\usepackage{multirow}
\usepackage{natbib}  
\usepackage{newfloat}
\usepackage{threeparttable}
\usepackage{times}  
\usepackage{xcolor}
\usepackage{xspace}
\usepackage{algorithm}
\usepackage{algorithmic}

\usepackage{newfloat}
\usepackage{listings}
\DeclareCaptionStyle{ruled}{labelfont=normalfont,labelsep=colon,strut=off} 
\floatstyle{ruled}
\newfloat{listing}{tb}{lst}{}
\floatname{listing}{Listing}
\title{FairGP: A Scalable and Fair Graph Transformer Using Graph Partitioning}
\author{
    Renqiang Luo\textsuperscript{\rm 1}, Huafei Huang\textsuperscript{\rm 2}, Ivan Lee\textsuperscript{\rm 2}, Chengpei Xu\textsuperscript{\rm 3}, Jianzhong Qi\textsuperscript{\rm 4}, Feng Xia\textsuperscript{\rm 5}\thanks{Corresponding Author}
}
\affiliations{
    \textsuperscript{\rm 1}Dalian University of Technology\\
    \textsuperscript{\rm 2}University of South Australia\\    
    \textsuperscript{\rm 3}The University of New South Wales\\
    \textsuperscript{\rm 4}The University of Melbourne\\
    \textsuperscript{\rm 5}RMIT University\\
    \{lrenqiang, hhuafei\}@outlook.com, Ivan.Lee@unisa.edu.au, \\
    chengpei.xu@unsw.edu.au, jianzhong.qi@unimelb.edu.au, f.xia@ieee.org
}

\usepackage{bibentry}

\begin{document}

\maketitle

\begin{abstract}
Recent studies have highlighted significant fairness issues in Graph Transformer (GT) models, particularly against subgroups defined by sensitive features.
Additionally, GTs are computationally intensive and memory-demanding, limiting their application to large-scale graphs. 
Our experiments demonstrate that graph partitioning can enhance the fairness of GT models while reducing computational complexity. 
To understand this improvement, we conducted a theoretical investigation into the root causes of fairness issues in GT models. 
We found that the sensitive features of higher-order nodes disproportionately influence lower-order nodes, resulting in sensitive feature bias. 
We propose Fairness-aware scalable GT based on Graph Partitioning (FairGP), which partitions the graph to minimize the negative impact of higher-order nodes. 
By optimizing attention mechanisms, FairGP mitigates the bias introduced by global attention, thereby enhancing fairness. 
Extensive empirical evaluations on six real-world datasets validate the superior performance of FairGP in achieving fairness compared to state-of-the-art methods.
The codes are available at https://github.com/LuoRenqiang/FairGP.

\end{abstract}

%

\section{Introduction}
Despite the reported capabilities of Graph Transformer (GT) models in graph representation learning, recent studies have exposed fairness concerns associated with these models~\cite{luo2024fairgt}.
In critical applications such as medical diagnoses~\cite{pan2024towards}, loan approval~\cite{zhang2024learning}, and e-commerce~\cite{agrawal2024no}, features such as gender, race, age, and region are legally protected to prevent discrimination and are thus considered sensitive features~\cite{wang2024simfair}.
Unfortunately, GTs generate biased predictions that discriminate against specific subgroups characterized by these sensitive features.

In addition, GTs are computationally demanding and memory-intensive when applied on large-scale graphs due to the complexity of global attention mechanisms~\cite{dao2022flashattention, chen2023nagphormer}.
Consequently, training deep GTs on large-scale graphs can be resource-intensive, necessitating sophisticated techniques for partitioning nodes into smaller mini-batches to alleviate the computational overhead~\cite{wu2023sgformer}. 
Partitioning strategies impact the performance of GTs and reduce the time and space complexity~\cite{blondel2008fast, traag2019louvain}. 
Additionally, these strategies may also be subject to fairness issues.

\textit{Will GT with graph partitioning exhibit the same fairness issues as a vanilla GT?}
Various efforts have been devoted to developing fairness-aware algorithms, aiming to control the degree to which a model depends on sensitive features, measured by independence criteria such as statistical parity (measured by $\Delta_\text{SP}$) and equality opportunity (measured by $\Delta_\text{EO}$)~\cite{zhang2024adversarial}.
We calculate the fairness metrics for both vanilla GT and GT with graph partitioning separately, as shown in Figure~\ref{fig:background}.
The experimental settings are detailed in Appendix A.1, along with results from three additional graph partitioning strategies~\cite{luo2024fairgp}.
The results show that graph partitioning improves fairness in GT models, indicating its potential to address fairness issues within GTs.
This observation raises a natural and fundamental question: \textit{how does graph partitioning promote fairness? }
To answer the question, we conducted a theoretical investigation into the underlying mechanisms.

\begin{figure}[H]
    \centering
	\includegraphics[width=0.4\textwidth]{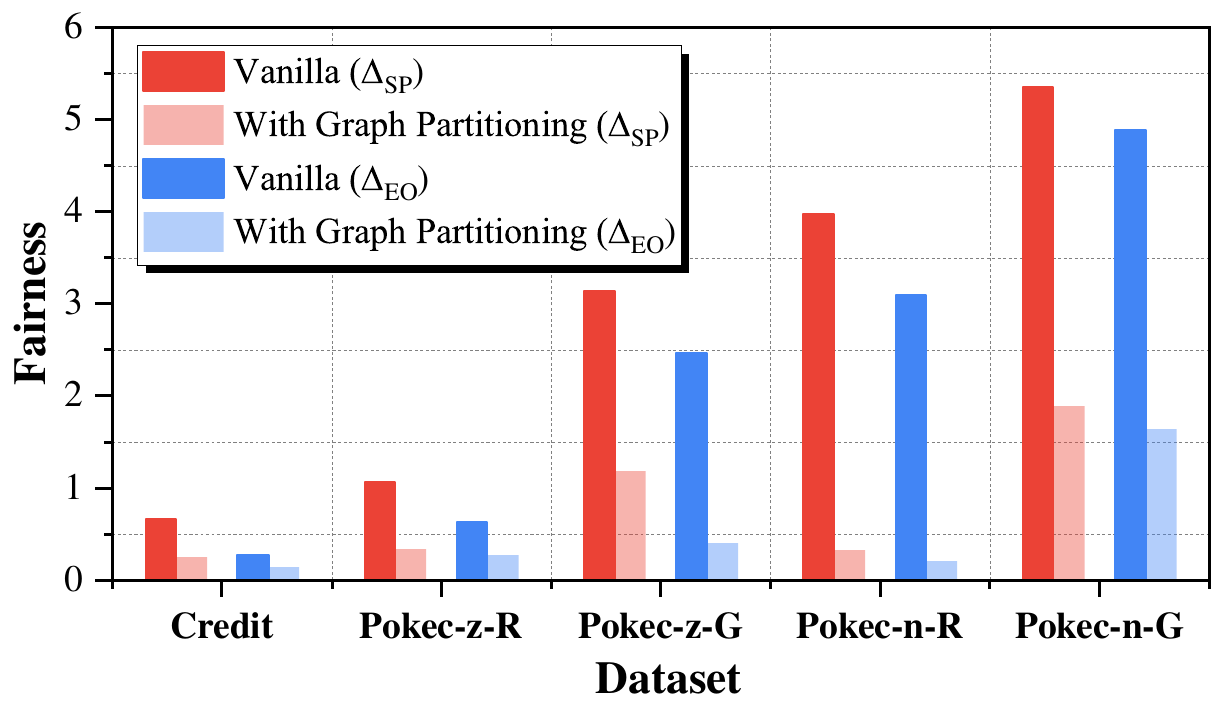}
    \caption{Graph partitioning increases GT fairness.}
    \label{fig:background}
\end{figure}

Intuitively, the global attention mechanism in GTs tends to favour higher-order nodes, causing the sensitive features of these nodes to disproportionately influence lower-order nodes, resulting in sensitive feature bias~\cite{shehzad2024graph}.
Our theoretical analysis indicates that global attention among nodes with different sensitive features significantly impacts GT fairness.
Minimizing the influence of higher-order nodes on nodes with varying sensitive features can effectively improve GT fairness.
In addition, we categorize the attention into inter-cluster and intra-cluster attention. 
To enhance fairness, we optimize both inter-cluster and intra-cluster attention, thereby reducing interactions between nodes with different sensitive features. 
By adapting the graph partitioning, we enable a large number of lower-order nodes to avoid the undue influence from higher-order nodes.
This effectively addresses the fairness issue, ensuring that the sensitive features of lower-order nodes are less affected by those of higher-order.

Building on this insight, we propose \textbf{Fair}ness-aware scalable GT based on \textbf{G}raph \textbf{P}artitioning (\textbf{FairGP}).
FairGP reduces the influence range of higher-order nodes in global attention by partitioning the graph and focusing attention within respective subgraphs. 
Additionally, FairGP optimizes attention to enhance the similarity between the distribution of the original sensitive features and the distribution of their embeddings. 
This optimization minimizes the negative influence between nodes with different sensitive features, thereby enhancing GT fairness. 
The main contributions of this work are summarized as follows:
\begin{itemize}
    \item We show that GT fairness is influenced by the attention of higher-order nodes and the attention between nodes with different sensitive features. 
    To the best of our knowledge, this work is the first to uncover the underlying cause of fairness issues in GTs.
    \item We propose a novel fairness-aware scalable GT, FairGP, which partitions the graph to mitigate the influence of higher-order node attention and optimizes inter-cluster attention to mitigate bias in nodes with different sensitive features. 
    \item To verify the effectiveness of the proposed FairGP, we conduct extensive empirical experiments comparing with state-of-the-art models. 
    The results show that FairGP consistently improves fairness by at least 40\% across both statistical parity and equality opportunity over six real-world datasets.
\end{itemize}

\section{Related Work}

\subsection{Fairness-aware Graph Neural Network (GNNs)}
Fairness in GNNs has gained substantial attention, particularly in efforts to identify and mitigate biases associated with specific sensitive features~\cite{zhang2024endowing}. 
Various fairness-aware GNN studies aim to preserve the independence of sensitive features through pre-processing and in-processing techniques~\cite{dong2023fairness, luo2024algorithmic}.

Graphair~\cite{ling2023learning} is a recent pre-processing algorithm. 
It automatically identifies fairness-aware augmentations from input graph data, aiming to circumvent sensitive features while preserving the integrity of other valuable information. 
FairAC~\cite{guo2023fair} proposes a fairness feature completion approach.
It completes missing information by adversarial learning and achieves fair node embeddings for graphs lacking certain features.
Unlike GNN methods, GTs that depend on the attention score matrix face challenges in directly identifying sensitive features in the input, making it difficult to apply pre-processing techniques effectively.
In-processing methods, on the other hand, focus on message-passing or convolution propagation. 
FairSIN~\cite{yang2024fairsin} introduces a neutralization-based paradigm, which incorporates additional fairness-enhancing features into node features or representations before message-passing. 
FMP~\cite{jiang2024chasing} directly addresses the use of sensitive features during forward propagation for node classification tasks using cross-entropy loss, eliminating the need for data pre-processing. 
However, the global attention mechanism in GTs, which facilitates direct node-to-node interactions, differs from the message-passing or convolution used in GNNs~\cite{yin2023lgi}.
In-processing techniques may not be directly applicable to GTs.

\subsection{Scalable Graph Transformer}
Recent GT studies are shifting their focus to handling the large-scale graph~\cite{xing2024less}.
DIFFormer~\cite{wu2023difformer} introduces a scalable GT framework guided by data-invariant functions, enabling efficient processing of large-scale graphs. 
Similarly, SGFormer~\cite{wu2023sgformer} simplifies the Transformer architecture to boost computational efficiency on large-scale graphs. 
Polynormer~\cite{deng2024polynormer} advances the field by developing a polynomial expressive GT that can capture features in complex graph structures more effectively.

Despite these advancements, existing GT methods often lack explicit consideration of algorithmic fairness. 
When applied to real-world datasets, they exhibit significant fairness issues.
FairGT~\cite{luo2024fairgt} addresses the fairness problem within GT based on a sensitive feature complete graph.
It faces challenges in deployment on large graphs due to its reliance on global attention in two whole graphs (original and sensitive feature complete graph).
Scalable techniques such as mini-batch and graph partitioning are challenging to deploy in FairGT while ensuring guaranteed performance, because the sensitive feature complete graph relies on all edges.
Addressing the fairness issues of GTs on large-scale graphs remains a pressing and current challenge.

\section{Preliminaries}

\subsection{Notations}
Unless otherwise specified, we denote sets with copperplate uppercase letters (e.g., $\mathcal{A}$), matrices with bold uppercase letters (e.g., $\mathbf{A}$), and vectors with bold lowercase letters (e.g., $\mathbf{x}$).
We denote the number of elements in the set $\mathcal{A}$ as $|\mathcal{A}|$.
We denote a graph as $\mathcal{G} = \{\mathcal{V}, \mathbf{A}, \mathbf{H}\}$, where $\mathcal{V}$ is the set of nodes in the graph and $|\mathcal{V}| = n$, $\mathbf{A} \in \mathbb{R}^{n \times n}$ is the adjacency matrix, and $\mathbf{H} \in \mathbb{R}^{n \times d}$ is the node feature matrix.
We follow the convention of NumPy in Python for matrix and vector indexing:
$\mathbf{A}[i,j]$ represents the entry of matrix $\mathbf{A}$ at the $i$-th row and the $j$-th column;
$\mathbf{A}[i,:]$ and $\mathbf{A}[:,j]$ represent the $i$-th row and the $j$-th column of matrix $\mathbf{A}$, respectively.
$\mathbf{H}[:,s]$ denotes a sensitive feature of nodes, and $\mathbf{H}[i,s]$ denotes a sensitive feature of node $v_i$.

\subsection{Fairness Evaluation Metrics}
We present two definitions of fairness for a binary ground truth label $y \in \{0,1\}$ and a binary sensitive feature $\mathbf{H}[i,s] \in \{0,1\}$.
We use $\hat{y} \in \{0,1\}$ to represent the predicted label.

\textit{Definition 1. Statistical Parity (i.e., Demographic Parity, Independence) }~\cite{dwork2012fairness}. 
Statistical parity requires the predictions to be independent of the sensitive features. 
The prediction can be formally written as:
\begin{equation}
    \mathbb{P}(\hat{y} = 1|\mathbf{H}[i,s] = 0)=\mathbb{P}(\hat{y} = 1|\mathbf{H}[i,s] = 1).
\label{equ:SP}
\end{equation}

When both the predicted labels and the sensitive features are binary, the extent of statistical parity can be quantified by $\Delta_\text{SP}$, defined as follows:
\begin{equation}
    \Delta_\text{SP}=|\mathbb{P}(\hat{y} = 1|\mathbf{H}[i,s] = 0)-\mathbb{P}(\hat{y} = 1|\mathbf{H}[i,s] = 1)|.
\label{equ:delta_SP}
\end{equation}

The $\Delta_\text{SP}$ measures the acceptance rate difference between two sensitive subgroups with $\mathbf{H}[i,s] = 0$ and $\mathbf{H}[i,s] = 1$.

\textit{Definition 2. Equal Opportunity}~\cite{hardt2016equality}. 
Equal opportunity requires that the probability of a positive instance leading to a positive outcome be the same across all subgroups. 
For individuals with positive ground truth labels, positive predictions should not depend on sensitive features. This can be formulated as follows:
\begin{equation}
    \begin{aligned}
        &\mathbb{P}(\hat{y} = 1|y = 1,\mathbf{H}[i,s] = 0) \\
        = &\mathbb{P}(\hat{y} = 1|y = 1,\mathbf{H}[i,s] = 1).
    \end{aligned}
\label{equ:EO}
\end{equation}

Fairness-aware algorithms prevent the allocation of unfavourable predictions to individuals solely based on their sensitive subgroup affiliation.
In particular, $\Delta_\text{EO}$ quantifies the extent of deviation in predictions from equal opportunity.
To quantitatively assess equal opportunity:
\begin{equation}
    \begin{aligned}
        \Delta_\text{EO}=|&\mathbb{P}(\hat{y} = 1|y = 1,\mathbf{H}[i,s] = 0) \\
        - &\mathbb{P}(\hat{y} = 1|y = 1,\mathbf{H}[i,s] = 1)|.    
    \end{aligned}
\label{equ:delta_EO}
\end{equation}

Both $\Delta_\text{SP}$ and $\Delta_\text{EO}$ are evaluated on the test set.

\subsection{Transformer}
The Transformer architecture consists of a composition of Transformer layers.
Each Transformer layer has two parts: a self-attention module and a position-wise feed-forward network (FFN).
Let $\mathbf{X} = [\mathbf{h}_1^\top, ... , \mathbf{h}_i^\top]^\top \in \mathbb{R}^{i \times d_h}$ be the input of the self-attention module where $d_h$ is the hidden dimension and $\mathbf{h}_j \in \mathbb{R}^{1 \times d_h}$ is the hidden representation at position $j$.
The input $\mathbf{X}$ is projected by three matrices $\mathbf{W}_\text{Q} \in \mathbb{R}^{d_h \times d_K}$, $\mathbf{W}_\text{K} \in \mathbb{R}^{d_h \times d_K}$, and $\mathbf{W}_\text{V} \in \mathbb{R}^{d_h \times d_V}$ to the corresponding representations Q, K, V.
The self-attention is then calculated as:
\begin{equation}
    \begin{aligned}
        &\text{Attn}(\mathbf{X})=\text{Softmax}\left(\frac{\text{Q}\text{K}^\top}{\sqrt{d_K}}\right)\text{V}, \\
        &\text{Q} = \mathbf{X}\mathbf{W}_\text{Q}, \text{K} = \mathbf{X}\mathbf{W}_\text{K}, V = \mathbf{X}\mathbf{W}_V.
    \end{aligned}
\end{equation}
To simplify, we consider single-head self-attention and assume $d_K=d_V=d_h$.
The extension to multi-head attention is straightforward, and we omit the bias terms for simplicity.
The attention score matrix is $\hat{\mathbf{A}} \in \mathbb{R}^{n \times n}$: 

\begin{equation}
    \begin{aligned}
        \hat{\mathbf{A}} = \text{Softmax}\left(\frac{\text{Q}\text{K}^\top}{\sqrt{d_K}}\right).    
    \end{aligned}
\end{equation}

$\hat{\mathbf{A}}[u,v]$ represents the attention score between nodes $u$ and $v$. 
Due to the Softmax function, we have $\sum^n_{i=1}\hat{\mathbf{A}}[i,:] = 1$ and $\sum^n_{i=1}\hat{\mathbf{A}}[:,j] = 1$.

\section{The Fairness Issues in GT}
In this section, we analyze the underlying cause of fairness issues in GTs.
Firstly, we present our empirical observations that the global attention mechanism negatively impacts fairness in GT.
Because, global attention tends to focus on the proportions of sensitive features among higher-order nodes, while fairness-aware algorithms often protect the proportions of sensitive features among all nodes (most of the nodes are lower-order nodes).
Methods that can truncate the attention score matrix, such as graph partitioning, can effectively mitigate this negative impact. 
Then, we present the theoretical findings on GT fairness and explain how graph partitioning enhances fairness. 
A theoretical foundation is provided, enabling us to propose a GT fairness algorithm (FairGP) based on graph partitioning.
\begin{table}[H]
  \centering
  \footnotesize
  \tabcolsep=0.1cm
  \begin{tabular}{lcccccc}
  \toprule
    \multirow{2}[2]{*}{\textbf{Datasets}} & \multicolumn{2}{c}{\textbf{All Nodes}} & \multicolumn{2}{c}{\textbf{Higher-order}} & \multicolumn{2}{c}{\textbf{Prediction ($\hat{y}=1$)}}\\
    \cline{2-7}
    & $s = 1$ & $s = 0$ & $s = 1$ & $s = 0$ & $s = 1$ & $s = 0$ \\
	\midrule
    \textbf{Credit} & \textbf{\textcolor{red}{10.17}} & 1 & \textbf{\textcolor{red}{2.25}} & 1 & \textbf{\textcolor{red}{2.94}} & 1 \\
    \textbf{Pokec-z-R} & \textbf{\textcolor{red}{1.82}} & 1 & 1 & \textbf{\textcolor{blue}{1.29}} & 1 & \textbf{\textcolor{blue}{2.63}} \\
    \textbf{Pokec-z-G} & 1 & 1 & 1 & \textbf{\textcolor{blue}{1.25}} & 1 & \textbf{\textcolor{blue}{3.10}} \\
    \textbf{Pokec-n-R} & \textbf{\textcolor{red}{2.12}} & 1 & \textbf{\textcolor{red}{2.51}} & 1 & \textbf{\textcolor{red}{3.50}} & 1 \\
    \textbf{Pokec-n-G} & \textbf{\textcolor{red}{1.26}} & 1 & 1 & \textbf{\textcolor{blue}{1.37}} & 1 & \textbf{\textcolor{blue}{2.69}} \\
    \textbf{AMine-L} & \textbf{\textcolor{red}{1.18}} & 1 & 1 & \textbf{\textcolor{blue}{1.36}} & 1  & \textbf{\textcolor{blue}{6.81}} \\
  \bottomrule
  \end{tabular}
  \caption{Distribution of different sensitive features. We denotes $\textbf{H}[i,s]=1$ as $s = 1$, and $\textbf{H}[i,s]=0$ as $s = 0$.}
  \label{tab:empirical_ovservation}
\end{table}

\subsection{Empirical Observations}
We empirically show how the global attention mechanism in GT tends to favour higher-order nodes.
The experimental settings are detailed in Appendix A.2.
We focus on the proportion of different sensitive features across different nodes, analyzing how these distributions are affected by GT training. 
This analysis includes the proportion of different sensitive features among all nodes, higher-order nodes, and prediction results.
We normalize the minority subgroups as $1$.
Thus, if $|\mathbf{H}[i,s] = 1| > |\mathbf{H}[i,s] = 0|$, for $i \in \{1,2, \dots, n\}$:
\begin{equation}
    \begin{aligned}
        &\textbf{All Nodes}(\mathbf{H}[i,s] = 1) = \frac{|\mathbf{H}[i,s] = 1|}{|\mathbf{H}[i,s] = 0|}, \\
        &\textbf{All Nodes}(\mathbf{H}[i,s] = 0) = 1,
    \end{aligned}
\end{equation}
and, if $|\mathbf{H}[i,s] = 0| > |\mathbf{H}[i,s] = 1|$:
\begin{equation}
    \begin{aligned}
        &\textbf{All Nodes}(\mathbf{H}[i,s] = 1) = 1, \\
        &\textbf{All Nodes}(\mathbf{H}[i,s] = 0) = \frac{|\mathbf{H}[i,s] = 0|}{|\mathbf{H}[i,s] = 1|}.
    \end{aligned}
\end{equation}

The definition is similar in higher-order nodes and prediction results, and the details are shown in Appendix B.

In Table~\ref{tab:empirical_ovservation}, higher proportions of sensitive features will be highlighted in red ($\mathbf{H}[i,s] = 1$) and blue ($\mathbf{H}[i,s] = 0$) for clarity.
When the overall distribution of sensitive features among all nodes differs from that of higher-order nodes, GTs tend to learn to predict based on the distribution of higher-order nodes.
It may even lead to the opposite prediction.
The majority population in both distributions is the same, the proportion of sensitive features in the prediction result will align more closely with the higher-order nodes.
Consequently, the proportions of sensitive features among higher-order nodes can affect the GT fairness.

\subsection{Theoretical Findings}
We denote the similarity between the distribution of original sensitive features and the distribution of sensitive features when they are mapped to node embeddings as sensitive feature similarity.
Node embeddings are computed by the global attention mechanism.
We select the Euclidean Norm to measure similarity. 
A higher similarity indicates greater independence of sensitive attributes during training, aligning with algorithmic fairness.
This analysis helps us understand how the attention mechanism impacts the preservation or alteration of sensitive feature distributions during the embedding process. 
By comparing these distributions, we can assess whether a GT model introduces or mitigates bias, thereby uncovering the underlying causes of fairness issues.

\textbf{Theorem 1}
\textit{The sensitive feature similarity is bounded by the attention scores between nodes with different sensitive features.}
\begin{equation}
    \Vert \mathbf{H}[:,s] - \hat{\textbf{A}}\mathbf{H}[:,s]\Vert_2 \leq \sum_{u, v \in \mathcal{V}} \sum_{\mathbf{H}[u, s] \neq \mathbf{H}[v, s]} \hat{\mathbf{A}}[u,v]
\end{equation}

\textit{Proof.}
Based on the Triangle Inequality:
\begin{equation*}
    \begin{aligned}
    &\Vert \mathbf{H}[:,s] - \hat{\textbf{A}}\mathbf{H}[:,s]\Vert_2 \\
    = &\sqrt{\sum_{u \in \mathcal{V}}(\mathbf{H}[u, s] - \sum_{v \in \mathcal{V}} \hat{\mathbf{A}}[u,v] \mathbf{H}[v, s])^2} \\
    \leq &\sum_{u \in \mathcal{V}} \sqrt{(\mathbf{H}[u, s] - \sum_{v \in \mathcal{V}} \hat{\mathbf{A}}[u,v] \mathbf{H}[v, s])^2} \\ 
    = &\sum_{u \in \mathcal{V}} \Vert \mathbf{H}[u, s] - \sum_{v \in \mathcal{V}} \hat{\mathbf{A}}[u,v] \mathbf{H}[v, s] \Vert_2 \\
    = &\sum_{u \in \mathcal{V}} \Vert \sum_{v \in \mathcal{V}} \hat{\mathbf{A}}[u,v] (\mathbf{H}[u, s] - \mathbf{H}[v, s]) \Vert_2 \\
    \leq & \sum_{u, v \in \mathcal{V}} \hat{\mathbf{A}}[u,v] \Vert \mathbf{H}[u, s] - \mathbf{H}[v, s] \Vert_2 \\
    = & \sum_{u, v \in \mathcal{V}} \sum_{\mathbf{H}[u, s] \neq \mathbf{H}[v, s]} \hat{\mathbf{A}}[u,v].
    \end{aligned}    
\end{equation*}
\rightline{$\square$}

Since global attention tends to concentrate on higher-order nodes~\cite{xing2024less}, algorithmic fairness is predominantly influenced by these higher-order nodes with different sensitive features.

Next, we consider GT with partitioning.
We denote $\hat{\mathbf{A}}'$ as the GT attention score matrix with graph partitioning.
We assume even partitions, and the number of partitions (i.e. clusters) is $c$.
Let $\alpha_{pq}$ represent the attention score between clusters $p$ and $q$.
Then the approximate attention score between node $u$ and $v$ (in different clusters) with graph partitioning can be expressed as $\hat{\mathbf{A}}'[u,v] = \frac{\alpha_{pq}}{n/c}$~\cite{xing2024less}.
Thus,
\begin{equation}
    \hat{\mathbf{A}}'[u,v] = \left\{ 
    \begin{aligned}
        &\hat{\mathbf{A}}[u,v], \quad &u \in \mathcal{V}_p, v \in \mathcal{V}_p \\
        &\frac{\alpha_{pq}}{n/c}, &u \in \mathcal{V}_p, v \in \mathcal{V}_q
    \end{aligned}
    \right.
\end{equation}

\textbf{Lemma 1} The sensitive feature similarity are lower than $\sqrt{n}$, whether using graph partitioning.
\begin{equation}
    \begin{aligned}
         &\Vert \mathbf{H}[:,s] - \hat{\textbf{A}}\mathbf{H}[:,s]\Vert_2 \leq \sqrt{n}, \\
         &\Vert \mathbf{H}[:,s] - \hat{\textbf{A}}'\mathbf{H}[:,s]\Vert_2 \leq  \sqrt{n}.
    \end{aligned}
\end{equation}

The proof of \textbf{Lemma 1} is shown in Appendix C.

\textbf{Theorem 2}
\textit{The change in sensitive feature similarity in GT with graph partitioning is constrained by the attention mechanism between different clusters.}
\begin{equation}
    \begin{aligned}
    &\big|\Vert \mathbf{H}[:,s] - \hat{\textbf{A}}\mathbf{H}[:,s]\Vert_2 - \Vert \mathbf{H}[:,s] - \hat{\textbf{A}}'\mathbf{H}[:,s]\Vert_2\big| \\
    \geq &\frac{1}{2 \sqrt{n}}|\sum_{u \in \mathcal{V}_p} (\sum_{v \in \mathcal{V}_q} \hat{\mathbf{A}}[u,v] (\mathbf{H}[u,s] - \mathbf{H}[v,s])^2 \\ 
    - &\sum_{u \in \mathcal{V}_p} (\sum_{v \in \mathcal{V}_q} \hat{\mathbf{A}}'[u,v] (\mathbf{H}[u,s] - \mathbf{H}[v,s])^2|,
    \end{aligned}    
\end{equation}
\textit{where $p \neq q$}.

\textit{Proof.}
Based on \textbf{Lemma 1},
\begin{equation*}
     \Vert \mathbf{H}[:,s] - \hat{\textbf{A}}\mathbf{H}[:,s]\Vert_2 + \Vert \mathbf{H}[:,s] - \hat{\textbf{A}}'\mathbf{H}[:,s]\Vert_2 \leq 2 \sqrt{n}.
\end{equation*}

In addition,
\begin{equation*}
    \Vert \mathbf{H}[:,s] - \hat{\textbf{A}}\mathbf{H}[:,s]\Vert_2 \geq 0, \Vert \mathbf{H}[:,s] - \hat{\textbf{A}}'\mathbf{H}[:,s]\Vert_2 \geq 0.
\end{equation*}

Thus,
\begin{equation*}
    \begin{aligned}
        &\big|\Vert \mathbf{H}[:,s] - \hat{\textbf{A}}\mathbf{H}[:,s]\Vert_2 - \Vert \mathbf{H}[:,s] - \hat{\textbf{A}}'\mathbf{H}[:,s]\Vert_2\big| \\
        = &\frac{\big|\Vert \mathbf{H}[:,s] - \hat{\textbf{A}}\mathbf{H}[:,s]\Vert^2_2 - \Vert \mathbf{H}[:,s] - \hat{\textbf{A}}'\mathbf{H}[:,s]\Vert^2_2\big|}{\Vert \mathbf{H}[:,s] - \hat{\textbf{A}}\mathbf{H}[:,s]\Vert_2 + \Vert \mathbf{H}[:,s] - \hat{\textbf{A}}'\mathbf{H}[:,s]\Vert_2} \\
        \geq &\frac{1}{2\sqrt{n}}\big| \Vert \mathbf{H}[:,s] - \hat{\textbf{A}}\mathbf{H}[:,s]\Vert^2_2 - \Vert \mathbf{H}[:,s] - \hat{\textbf{A}}'\mathbf{H}[:,s]\Vert^2_2\big| \\
        = &\frac{1}{2\sqrt{n}} \left|\Big(\sum_{u \in \mathcal{V}} \big ( \sum_{v \in \mathcal{V}} \hat{\mathbf{A}}[u,v]( \mathbf{H}[u, s] - \mathbf{H}[v, s])\big)^2 \right.\\
        - &\left. \sum_{u \in \mathcal{V}} \big( \sum_{v \in \mathcal{V}} \hat{\mathbf{A}}'[u,v]( \mathbf{H}[u, s] - \mathbf{H}[v, s])\big)^2 \Big)\right|. \\
    \end{aligned}
\end{equation*}
\begin{figure*}[htbp]
	\centering
    \includegraphics[width=0.85\textwidth]{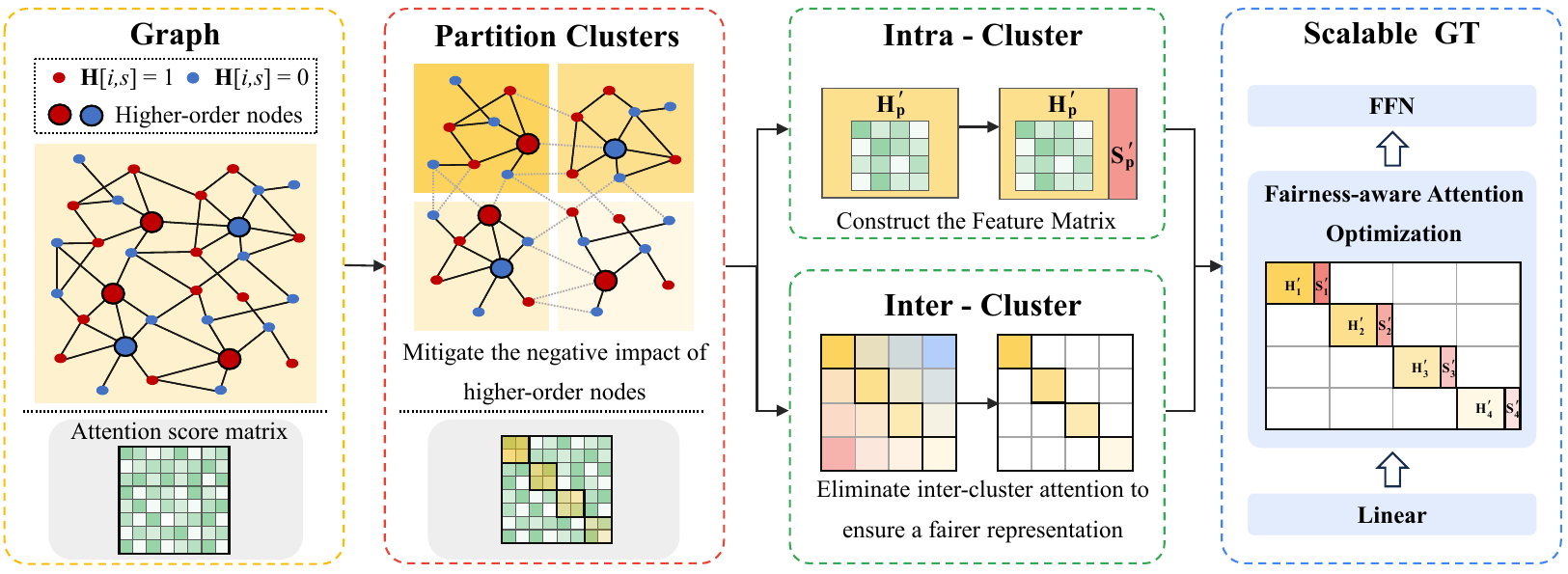}
    \caption{The illustration of FairGP.}
    \label{fig:architecture}
\end{figure*}

Because, 
\begin{equation*}
    \begin{aligned}
        \sum_{u \in \mathcal{V}_p} \sum_{v \notin \mathcal{V}_p} \hat{\mathbf{A}}[u,v]( \mathbf{H}[u, s] - \mathbf{H}[v, s]) = 0, \\
        \sum_{u \in \mathcal{V}_p} \sum_{v \notin \mathcal{V}_p} \hat{\mathbf{A}}'[u,v]( \mathbf{H}[u, s] - \mathbf{H}[v, s]) = 0.
    \end{aligned}
\end{equation*}

In addition, based on equation (10), if $u \in \mathcal{V}_p,v \in \mathcal{V}_p$:
\begin{equation*}
    \hat{\mathbf{A}}[u,v] = \hat{\mathbf{A}}'[u,v].    
\end{equation*}

Thus,
\begin{equation*}
    \begin{aligned}
        &\left|\Vert \mathbf{H}[:,s] - \hat{\textbf{A}}\mathbf{H}[:,s]\Vert_2 - \Vert \mathbf{H}[:,s] - \hat{\textbf{A}}'\mathbf{H}[:,s]\Vert_2\right| \\
        \geq &\frac{1}{2\sqrt{n}} \left|\Big(\sum_{u \in \mathcal{V}_p} \big( \sum_{v \in \mathcal{V}_p} \hat{\mathbf{A}}[u,v]( \mathbf{H}[u, s] - \mathbf{H}[v, s]) \right.\\
        + &\sum_{v \notin \mathcal{V}_p} \hat{\mathbf{A}}[u,v]( \mathbf{H}[u, s] - \mathbf{H}[v, s])\big)^2 \\
        - &\big(\sum_{u \in \mathcal{V}_p} ( \sum_{v \in \mathcal{V}_p} \hat{\mathbf{A}}'[u,v]( \mathbf{H}[u, s] - \mathbf{H}[v, s]) \\
        + &\left.\sum_{v \notin \mathcal{V}_p} \hat{\mathbf{A}}'[u,v]( \mathbf{H}[u, s] - \mathbf{H}[v, s])\big)^2\Big) \right|\\
        = &\frac{1}{2 \sqrt{n}}\left|\sum_{u \in \mathcal{V}_p} \big(\sum_{v \in \mathcal{V}_q} \hat{\mathbf{A}}[u,v](\mathbf{H}[u,s] - \mathbf{H}[v,s])\big)^2 \right.\\ 
        - &\left. \sum_{u \in \mathcal{V}_p} \big(\sum_{v \in \mathcal{V}_q} \hat{\mathbf{A}}'[u,v] (\mathbf{H}[u,s] - \mathbf{H}[v,s])\big)^2\right|.
    \end{aligned}
\end{equation*}
\rightline{$\square$}

The sensitive feature similarity bound is maximized when $\sum_{u \in \mathcal{V}_p} \sum_{v \in \mathcal{V}_q}\hat{\mathbf{A}}'[u,v] = 0$ ($p \neq q$). 
Thus, setting inter-cluster attention scores to zero enhances GT fairness.

\section{FairGP}
We are now ready to present our model FairGP to enhance GT fairness with graph partitioning. 
As shown in Figure~\ref{fig:architecture}, we first encode the structural topology by constructing the node feature matrix. 
Next, we partition the graph into different clusters, which helps reduce attention computation time complexity and enhance fairness. 
Finally, we mitigate inter-cluster attention to address fairness issues arising from the higher-order node bias.
\begin{table*}[h]
    \centering
    \scriptsize
    \tabcolsep=0.15cm
		\begin{tabular}{lcccccccc}
			\toprule
			\multirow{2}[2]{*}{\textbf{Methods}} & \multicolumn{4}{c}{Pokec-z-R} & \multicolumn{4}{c}{Pokec-z-G} \\
            \cline{2-9}
			& ACC(\%) $\uparrow$ & AUC(\%) $\uparrow$  & $\Delta_\text{SP}$(\%) $\downarrow$   & $\Delta_\text{EO}$(\%) $\downarrow$  & ACC(\%) $\uparrow$ & AUC(\%) $\uparrow$ & $\Delta_\text{SP}$(\%) $\downarrow$   & $\Delta_\text{EO}$(\%) $\downarrow$ \\
			\midrule
			GCN & $66.24 \pm 2.12$    & $72.78 \pm 0.37$  & $7.32 \pm 1.48$  & $7.60 \pm 1.87$  & $65.09 \pm 0.92$  & $71.29 \pm 0.73$  & $6.68 \pm 2.57$   & $2.31 \pm 0.61$  \\
            GAT   & $66.01 \pm 0.69$  & $70.76 \pm 1.30$  & $4.71 \pm 1.05$  & $2.72 \pm 0.85$  & $66.49 \pm 0.23$  & $70.14 \pm 0.19$ & $11.21 \pm 1.56$  & $7.02 \pm 1.17$\\
            APPNP      & $65.24 \pm 1.26$  & $70.91 \pm 1.46$  & $4.52 \pm 1.02$  & $1.78 \pm 0.34$  & $64.91 \pm 0.60$  & $71.08 \pm 2.18$ & $6.17 \pm 2.40$  & $3.44 \pm 1.90$\\
			FairGNN      & $64.04 \pm 0.90$ & $71.75 \pm 1.65$  & $4.95 \pm 0.21$   & $4.29 \pm 0.20$  & $66.21 \pm 2.21$  & $70.12 \pm 0.39$  & $2.39 \pm 0.48$   & $2.62 \pm 0.32$\\
			FairSIN   & $67.76 \pm 0.71$ & $72.49 \pm 1.54$  & $1.49 \pm 0.74$   & \textcolor{blue}{\underline{$0.59 \pm 0.50$}}  & $65.91 \pm 1.04$  & $70.12 \pm 0.39$  & $2.00 \pm 2.49$   & $2.62 \pm 1.54$\\
            FMP     & $64.96 \pm 0.50$ & $72.42 \pm 0.40$  & $0.81 \pm 0.40$   & $1.73 \pm 1.03$  & $65.13 \pm 1.12$  & $70.21 \pm 1.82$  & $2.39 \pm 0.84$   & $2.26 \pm 0.23$\\
            FUGNN     & \textcolor{blue}{\underline{$68.38 \pm 0.43$}} & $72.53 \pm 1.13$  & \textcolor{blue}{\underline{$0.53 \pm 0.27$}}   & $1.32 \pm 0.95$  & $66.64 \pm 1.45$  & \textcolor{red}{$\mathbf{71.93 \pm 1.14}$}  & $3.83 \pm 0.81$   & $2.79 \pm 0.43$\\
            DIFFormer    & $66.42 \pm 1.18$ & \textcolor{blue}{\underline{$73.27 \pm 1.15$}} & $1.63 \pm 1.13$  & $1.94 \pm 0.46$ & $66.17 \pm 1.16$  & $71.39 \pm 2.26$  & $2.50 \pm 0.04$ & \textcolor{blue}{\underline{$1.64 \pm 0.40$}}\\
            SGFormer    & $65.43 \pm 0.11$ & $72.41 \pm 0.85$  & $2.32 \pm 0.06$   & $1.76 \pm 0.79$  & $66.29 \pm 0.70$  & \textcolor{blue}{\underline{$71.50 \pm 0.48$}}  & \textcolor{blue}{\underline{$1.91 \pm 1.02$}} & $2.33 \pm 0.82$\\
            Polynormer    & $65.22 \pm 1.35$ & $71.89 \pm 0.44$  & $2.34 \pm 0.06$   & $2.40 \pm 0.42$  & $65.82 \pm 0.58$  & $69.94 \pm 0.53$  & $5.49 \pm 0.41$   & $2.59 \pm 1.03$\\
            CoBFormer     & $66.72 \pm 1.04$ & $73.00 \pm 0.44$  & $2.45 \pm 0.49$   & $1.89 \pm 0.39$  & \textcolor{red}{$\mathbf{67.23 \pm 0.39}$}  & $70.57 \pm 0.18$  & $2.40 \pm 0.07$  & $2.38 \pm 1.32$\\   
            FairGT   & OOM & OOM & OOM & OOM & OOM & OOM & OOM & OOM\\
            \textbf{FairGP}  &  \textcolor{red}{$\mathbf{68.50 \pm 0.43}$} & \textcolor{red}{$\mathbf{73.83 \pm 0.25}$}  & \textcolor{red}{$\mathbf{0.33 \pm 0.12}$}   & \textcolor{red}{$\mathbf{0.27 \pm 0.16}$}  & \textcolor{blue}{\underline{$66.82 \pm 0.03$}}  & $71.00 \pm 0.47$  & \textcolor{red}{$\mathbf{1.18 \pm 0.24}$}   & \textcolor{red}{$\mathbf{0.40 \pm 0.24}$}\\
            \midrule
            \midrule
            \multirow{2}[2]{*}{\textbf{Methods}} & \multicolumn{4}{c}{Credit} & \multicolumn{4}{c}{AMiner-L}\\
            \cline{2-9}
			& ACC(\%) $\uparrow$ & AUC(\%) $\uparrow$  & $\Delta_\text{SP}$(\%) $\downarrow$   & $\Delta_\text{EO}$(\%) $\downarrow$  & ACC(\%) $\uparrow$ & AUC(\%) $\uparrow$ & $\Delta_\text{SP}$(\%) $\downarrow$   & $\Delta_\text{EO}$(\%) $\downarrow$ \\
			\midrule
			GCN  & $73.87 \pm 2.48$ & \textcolor{blue}{\underline{$61.45 \pm 2.34$}}  & $12.86 \pm 1.84$   & $10.63 \pm 0.24$  & $92.61 \pm 0.74$  & $90.87 \pm 1.19$  & $7.21 \pm 0.17$   & $4.62 \pm 1.14$\\
            GAT   & $68.29 \pm 1.84$ & $60.86 \pm 1.43$  & $9.74 \pm 2.04$   & $6.42 \pm 0.26$  & $ 90.31 \pm 0.41$  & $90.46 \pm 1.91$  & $6.57 \pm 0.20$   & $3.40 \pm 1.17$\\
            APPNP  & $74.19 \pm 0.79$ & $61.06 \pm 1.08$  & $13.30 \pm 0.94$   & $9.47 \pm 1.86$  & \textcolor{blue}{\underline{$92.75 \pm 1.29$}}  & $90.49 \pm 0.67$  & $7.00 \pm 0.07$   & $4.00 \pm 0.64$\\
			FairGNN & $68.29 \pm 2.25$ & \textcolor{red}{$\mathbf{61.73 \pm 1.39}$}  & $9.74 \pm 0.28$   & $8.83 \pm 0.46$  & $91.51 \pm 1.86$  & $90.16 \pm 0.29$  & $5.65 \pm 0.34$   & $3.78 \pm 0.32$\\
			FairSIN & \textcolor{blue}{\underline{$78.09 \pm 0.27$}} & $60.36 \pm 0.79$  & $1.85 \pm 0.42$   & $1.23 \pm 0.28$  & OOM  & OOM & OOM & OOM \\
            FMP & $74.51 \pm 1.41$ & $59.94 \pm 0.06$  & $2.31 \pm 0.44$   & $1.27 \pm 0.01$  & \textcolor{red}{$\mathbf{92.95 \pm 0.11}$}  & $91.11 \pm 1.11$  & $6.81 \pm 0.29$   & \textcolor{blue}{\underline{$3.23 \pm 0.78$}}\\
            FUGNN  & $77.02 \pm 0.07$ & $59.26 \pm 1.24$  & \textcolor{blue}{\underline{$0.62 \pm 0.48$}}   & \textcolor{blue}{\underline{$0.18 \pm 0.09$}}  & $90.85 \pm 1.70$  & $90.56 \pm 0.32$  & \textcolor{blue}{\underline{$4.57 \pm 0.60$}}   & $3.45 \pm 0.04$\\
            DIFFormer & $77.41 \pm 0.85$ & $60.94 \pm 1.77$  & $1.94 \pm 0.72$   & $2.47 \pm 0.15$  & $90.43 \pm 0.86$  & $90.42 \pm 0.15$  & $5.81 \pm 0.79$   & $3.60 \pm 0.83$\\
            SGFormer & $75.85 \pm 0.69$ & $60.95 \pm 0.87$  & $2.27 \pm 0.85$ & $2.32 \pm 0.27$  & $90.81 \pm 0.66$  & $90.02 \pm 0.66$  & $5.48 \pm 1.38$   & $3.27 \pm 0.27$\\
            Polynormer & $76.13 \pm 0.83$ & $59.95 \pm 2.61$  & $1.81 \pm 0.14$   & $1.67 \pm 0.10$  & $91.75 \pm 1.27$  & \textcolor{blue}{\underline{$91.69 \pm 0.13$}}  & $6.87 \pm 1.15$   & $3.98 \pm 0.60$\\
            CoBFormer & $76.40 \pm 0.21$ & $60.84 \pm 2.41$  & $2.92 \pm 0.96$   & $2.39 \pm 0.60$  & $90.00 \pm 1.73$  & $89.11 \pm 0.14$  & $5.00 \pm 1.08$   & $4.02 \pm 0.25$\\   
            FairGT & $77.85 \pm 0.64$ & $61.38 \pm 0.28$  & $1.89 \pm 0.59$   & $2.05 \pm 1.02$  & OOM & OOM & OOM & OOM \\
            \textbf{FairGP} & \textcolor{red}{$\mathbf{78.13 \pm 0.08}$} & $60.05 \pm 2.98$  & \textcolor{red}{$\mathbf{0.25 \pm 0.13}$} & \textcolor{red}{$\mathbf{0.14 \pm 0.03}$}  & $91.35 \pm 1.28$  & \textcolor{blue}{\underline{$91.12 \pm 1.19$}}  & \textcolor{red}{$\mathbf{4.44 \pm 0.13}$}   & \textcolor{red}{$\mathbf{2.97 \pm 0.37}$}\\
			\bottomrule
		\end{tabular}
	\caption{Comparison on utility (ACC and AUC) and fairness ( $\Delta_\text{SP}$ and $\Delta_\text{EO}$ ) in percentage (\%). $\uparrow$ denotes the larger, the better; $\downarrow$ denotes the opposite. The best results are red and bold, runner-ups are blue and underlined. OOM means out-of-memory.}
    \label{tab:result}
\end{table*}
\subsection{Constructing the Feature Matrix}
For the node feature matrix, We select the eigenvectors corresponding to the $t$ largest eigenvalues to construct the structure matrix $\mathbf{S} \in \mathbb{R}^{n \times t}$.
Then, We combine the original feature matrix $\mathbf{H}$ with the structure matrix $\mathbf{S}$ to preserve both node features and structural information:

\begin{equation}
\label{equ
information}
\mathbf{H}' = \mathbf{H} || \mathbf{S},
\end{equation}
where $||$ represents the concatenation operator, and $\mathbf{H}' \in \mathbb{R}^{n \times (d+t)}$ denotes the fused feature matrix. 
FairGT~\cite{luo2024fairgt} shows that the largest eigenvalues are closely tied to structural information, which tends to be fairer.
Thus, the feature matrix contributes to a fairer encoding.

\subsection{Graph Partitioning}
According to the empirical observations and \textbf{Theorem 1}, graph partitioning improves GT fairness by reducing the biased influence of higher-order nodes.
FairGP partitions the graph into $c$ non-overlapping clusters using METIS~\cite{karypis1998a}.
Let $\mathcal{G}_p = \{ \mathcal{V}_p, \mathcal{E}_p \}$ represents a subgraph of $\mathcal{G}$, satisfying $\bigcup^c_{p=1} \mathcal{G}_p = \mathcal{G}$ and $\bigcap^c_{p=1} \mathcal{G}_p = \emptyset$.

\subsection{Fairness-aware Attention Optimization}
GTs utilize global attention mechanism to capture information between node pairs, causing the higher-order node bias.
To obtain the updated hidden representations $\mathbf{I} \in \mathbb{R} ^ {n \times n}$.
\begin{equation}
    \begin{aligned}
        &\mathbf{I} = \text{FFN} \left(\text{Softmax}\left(\frac{\text{Q}\text{K}^\top}{\sqrt{n}}\right) \text{V}\right), \\
        &\text{Q} = \mathbf{H}' W^\text{Q}, \text{K} = \mathbf{H}' W^\text{K}, V = \mathbf{H}' W^\text{V},
    \end{aligned}
\end{equation}
where $W^{\text{Q}}$, $W^{\text{K}}$, and $W^{\text{V}} \in \mathbb{R} ^ {n \times n}$.
The $W^{\text{Q}}$, $W^{\text{K}}$, and $W^{\text{V}}$ are trainable weights of the linear layers in the Transformer, and FFN represents a Feed-Forward Neural Network.

FairGP separates the attention into inter-cluster and intra-cluster.
For the intra-cluster attention $\hat{\mathbf{A}}'[u,v]$ where $u,v \in \mathcal{V}_p$, the attention is the same as the whole attention score:
\begin{equation}
    \begin{aligned}
        \hat{\mathbf{A}}'[u,v] = \hat{\mathbf{A}}[u,v].
    \end{aligned}
\end{equation}

According to \textbf{Theorem 2}, the attention between different clusters is zero, which helps improve fairness.
To achieve this goal, we optimize the attention mechanism accordingly:
\begin{equation}
    \begin{aligned}
        \hat{\mathbf{A}}'[u,v] = 0, \quad u \in \mathcal{V}_p \quad \text{and} \quad v \in \mathcal{V}_q. 
    \end{aligned}
\end{equation}

This can nullify inter-cluster attention, enhancing GT fairness.
By focusing on intra-cluster attention and eliminating inter-cluster attention, FairGP ensures a fairer representation of the graph's structural information.

\subsection{Complexity of FairGP}
Dividing the graph into smaller clusters can reduce the computational burden while preserving essential structural information.
For each cluster, the time and space complexity is $O\big((\frac{n}{c})^2\big)$, making FairGP's overall complexity $O(\frac{1}{c^2}n^2)$.
This makes our approach scalable and efficient, ensuring that FairGP can handle large-scale graphs effectively.

\section{Experiments}
\subsection{Datasets and Implementation Details}
In our experiment, the task is node classification, tested on four datasets:  
\textbf{Credit}, \textbf{Pokec-z}, \textbf{Pokec-n}, and \textbf{AMiner-L}. 
Specifically, the datasets are labelled as \textbf{Pokec-z-R} and \textbf{Pokec-n-R} when living region is the sensitive feature, and \textbf{Pokec-z-G} and \textbf{Pokec-n-G} when gender is the sensitive feature.
More details are shown in Appendix D.

\subsection{Baselines}
We use three types of baseline: \textbf{GNNs} (GCN~\cite{kipf2017semi}, GAT~\cite{velickovic2018graph}, APPNP~\cite{klicpera2019predict}), \textbf{fairness-aware GNNs} (FairGNN~\cite{dai2023learning}, FairSIN~\cite{yang2024fairsin}, FMP~\cite{jiang2024chasing}, FUGNN~\cite{luo2024fugnn}), and \textbf{GTs} (DIFFormer~\cite{wu2023difformer}, SGFormer~\cite{wu2023sgformer}, Polynormer~\cite{deng2024polynormer}, CoBFormer~\cite{xing2024less}, FairGT~\cite{luo2024fairgt}).

\subsection{Multi-Class Sensitive Features}
There are multi-class sensitive features in dataset \textbf{AMiner-L}.
The previous fairness evaluation metrics are formulated for binary scenarios, but they can be easily extended into multi-class sensitive
feature scenarios following previous work~\cite{zhao2023fairness}. 
The main idea is to guarantee fairness across all pairs of sensitive subgroups. 
Quantification strategies can be applied by leveraging either the variance across different sensitive subgroups.
Specifically, the fairness evaluation metrics in multi-class sensitive features are defined as follows:
\begin{equation}
    \begin{aligned}        
    &\Delta_\text{SP}=\text{Var}_{i=1}^m\big(|\mathbb{P}(\hat{y}=1|s=i)\big), \\        &\Delta_\text{EO}=\text{Var}_{i=1}^m\big(|\mathbb{P}(\hat{y}=1|y=1,s=i)\big),
    \end{aligned}
\end{equation}
where class number denotes $m$.

\subsection{Comparison Results}
Table~\ref{tab:result} offers a detailed comparison of the fairness evaluation metrics for our proposed FairGP method against various baseline models across four real-world datasets.
Further comparison results for other datasets are included in Appendix E. 
Notably, \textbf{Pokec-z-R} and \textbf{Pokec-z-G} are the same datasets examined under two different sensitive features.
FairGP effectively addresses the multi-sensitive feature issue, as confirmed by the correlation results.
We present metrics such as overall Accuracy, AUC, $\Delta_{\text{SP}}$, and $\Delta_{\text{EO}}$. 
The table clearly indicates that FairGP consistently achieves superior fairness, validating the efficacy of our approach in fairness-aware node classification. 
FairGP not only improves fairness compared to traditional GTs but also surpasses existing fairness-aware GNN methods in fairness metrics. 
This highlights its effectiveness in addressing fairness-related issues in scalable GTs, demonstrating the practicality of FairGP for real-world applications.

\subsection{Ablation Study}
FairGP contains three key components: feature matrix, graph partitioning and attention optimization.
To assess their contributions, we conducted an ablation study, removing each component independently to evaluate its impact on prediction fairness.

\begin{table}[h]
  \centering
  \scriptsize
  \tabcolsep=0.05cm
  \renewcommand{\arraystretch}{1.5}
  \begin{tabular}{lccccc}
    \toprule
    \textbf{Dataset} & \textbf{Metric} & \textbf{FairGP} & \textbf{w/o FM} & \textbf{w/o GP} & \textbf{w/o AO}\\
    \midrule
    \textbf{Pokec-z-R} &  ACC $\uparrow$ & $\mathbf{68.50}_{\pm 2.01}$  & $68.16_{\pm 1.06}$ & $66.45_{\pm 1.98}$ & $68.20_{\pm 2.35}$ \\
    & AUC $\uparrow$ & $\mathbf{73.83}_{\pm 0.25}$ & $73.16_{\pm 1.23}$ & $71.97_{\pm 2.24}$ & $73.11_{\pm 1.16}$ \\
    & $\Delta_{\text{SP}}$ $\downarrow$ & $\mathbf{0.33}_{\pm 0.12}$ & $1.07_{\pm 0.31}$ & $2.14_{\pm 0.69}$ & $2.44_{\pm 1.36}$ \\
    & $\Delta_{\text{EO}}$ $\downarrow$ & $\mathbf{0.27}_{\pm 0.16}$ & $0.65_{\pm 0.39}$ & $5.44_{\pm 0.92}$ & $1.19_{\pm 0.61}$ \\
    \textbf{Pokec-z-G} & ACC $\uparrow$ & $\mathbf{66.82}_{\pm 0.03}$ & $66.21_{\pm 0.74}$ & $64.83_{\pm 1.90}$ & $66.52_{\pm 0.24}$ \\
    & AUC $\uparrow$ & $71.00_{\pm 0.47}$ & $70.57_{\pm 0.41}$ & $\mathbf{71.61}_{\pm 3.16}$ & $70.26_{\pm 0.80}$ \\
    & $\Delta_{\text{SP}}$ $\downarrow$ & $\mathbf{1.18}_{\pm 0.40}$ & $3.14_{\pm 0.31}$ & $5.84_{\pm 0.69}$ & $2.37_{\pm 0.59}$ \\
    & $\Delta_{\text{EO}}$ $\downarrow$ & $\mathbf{0.40}_{\pm 0.24}$ & $2.47_{\pm 0.90}$ & $4.99_{\pm 1.04}$ & $1.85_{\pm 0.39}$ \\
    \textbf{AMiner-L} & ACC $\uparrow$ & $\mathbf{91.35}_{\pm 1.28}$ & $91.03_{\pm 0.33}$ & $89.47_{\pm 0.70}$ & $90.47_{\pm 0.76}$ \\
    & AUC $\uparrow$ & $91.12_{\pm 1.19}$ & $91.21_{\pm 0.29}$ & $\mathbf{93.76}_{\pm 0.33}$ & $90.60_{\pm 0.80}$ \\
    & $\Delta_{\text{SP}}$ $\downarrow$ & $\mathbf{4.44}_{\pm 0.13}$ & $5.05_{\pm 0.21}$ & $5.95_{\pm 0.85}$ & $5.06_{\pm 0.12}$ \\
    & $\Delta_{\text{EO}}$ $\downarrow$ & $\mathbf{2.97}_{\pm 0.37}$ & $3.18_{\pm 0.58}$ & $3.32_{\pm 0.60}$ & $4.29_{\pm 1.13}$ \\
    \bottomrule
  \end{tabular}
  \caption{Ablation study of FairGP.}
  \label{tab:ablation}
\end{table}

The FairGP without graph partitioning is denoted as \textbf{w/o GP}.
Since GT without graph partitioning encounters out-of-memory issues in large-scale graphs, we employ mini-batch processing, a typical scalable technique, to keep it running.
Even when compared to results using the mini-batch technique, FairGP shows a significant improvement in fairness, with enhancements of at least 25\% across all datasets, and in some instances, exceeding 90\%.
These results validate our findings that graph partitioning improves GT fairness. 
 
The FairGP with a feature matrix that excludes structural features is denoted as \textbf{w/o FM}, and without attention optimization is denoted as \textbf{w/o AO}.
Notably, when comparing \textbf{w/o FM}, and \textbf{w/o AO}, FairGP consistently outperforms them in terms of accuracy and fairness.
The results shown in Table \ref{tab:ablation} reinforce the necessity of graph partitioning and attention optimization within FairGP.
More ablation experiments for other datasets are shown in Appendix F.

\subsection{Training Cost Comparison}
FairGP, partitioning a graph into clusters, exhibits efficiency compared to GTs.
To validate the efficiency of FairGP, we compare its training time with GT baselines.
For a fair comparison, we standardize key parameters across all methods,
setting the number of hidden dimensions to 128, the number
of layers to 1, and the number of heads to 1.
In addition, all models are evaluated with a runtime of 100 epochs, and the units are in seconds.
The runtimes of different sensitive features in the same dataset are identical. 
Therefore, we combine the results of \textbf{Pokec-z-R} and \textbf{Pokec-z-G} into \textbf{Pokec-z}.
The same approach is applied to \textbf{Pokec-n}.
The runtimes of FairGP and GT baselines are shown in Table~\ref{tab:runtime}, and the runtimes of computing $\mathbf{S}$ are shown in Appendix G.
\begin{table}[htbp]
	\centering
    \scriptsize
        \centering
        \tabcolsep=0.1cm
		\begin{tabular}{lccccc}
			\toprule
			& \textbf{DIFFormer} & \textbf{SGFormer} & \textbf{Polynormer} &\textbf{CoBFormer} &\textbf{FairGP} \\ 
            \midrule
            \textbf{Pokec-z}  & $138.73$ & $80.60$ & $18.34$ & $15.66$ & $6.88$ \\
            \textbf{Pokec-n} & $124.43$ & $79.39$ & $15.69$ & $11.38$ & $6.14$ \\
            \textbf{Aminer-L} & $830.44$ & $708.34$ & $70.03$ & $66.84$ & $32.77$ \\
            \bottomrule
		\end{tabular}
	\caption{Runtimes (s).}
\label{tab:runtime}
\end{table}
\section{Conclusion}
We showed how graph partitioning addresses GT fairness issues on large-scale graphs. 
We proposed FairGP, a novel algorithm to enhance scalable GT fairness. 
FairGP is built on graph partitioning, while graph structure information is encoded in the node features to enhance the sensitive feature similarity and the attention computation is adapted to decrease the negative impact of global attention. 
We also present an empirical study of the efficacy of the proposed approach. 
In diverse real-world datasets, FairGP earns the best $\Delta_\text{SP}$ and $\Delta_\text{EO}$ comparing state-of-the-art methods.
Moreover, in the vast majority of cases, FairGP not only improved fairness but also achieved leading performance, further demonstrating its effectiveness and superiority in large-scale graph applications.
In future work, we will further expand FairGP to address situations with limited sensitive features. 
Since FairGP relies on access to the entire graph for partitioning, it is less suited for applications where privacy preservation or significant misinformation is a concern. 
In the future, we aim to extend FairGP to be applicable in attribute-constrained and more complex scenarios.

\bibliography{ArXiv_Luo_FairGP}

\newpage
\section*{Appendix A.1}
\textbf{Experiment Settings}: All experiments are conducted on the Ubuntu operating system, using 2 V100 GPU with 16GB memory. 
The CUDA version 11.6, PyTorch 1.13.1, and Torch Geometric 2.5.3.
For hyperparameters, we search the hidden number from \{16, 32, 64, 128\}, the number of partition clusters from \{100, 150, 200, 250, 300\}, and choose the METIS algorithm~\cite{karypis1998a} for graph partition. 
For the training set size, we select 6,000 nodes for \textbf{Credit}, 5,000 nodes for \textbf{AMiner-L}, and 1,000 nodes for the remaining datasets. 
Note that Vanilla GT would cause an out-of-memory error. 
Therefore, we implemented it through a mini-batch strategy. 

\textbf{Impact of Graph Partitioning on Fairness}:  We also test the effect of different partitioning strategies on the GT fairness. 
Specifically, we select Louvain~\cite{blondel2008fast}, Leiden~\cite{traag2019louvain}, and Random partitioning. 
The hidden number is set to 64, the cluster number is set to 100, and the results are shown in Tables~\ref{tab:partition_mini_louvain},~\ref{tab:partition_mini_leiden}, and ~\ref{tab:partition_mini_random}. 
It can be seen that using partitioning strategies improves fairness performance in our experiments.

\begin{table}[H]
\centering
\footnotesize
\tabcolsep=0.15cm
\vspace{-1em}
\begin{tabular}{l|cc|cc}
\toprule
\multicolumn{1}{c}{\multirow{2}{*}{\textbf{Dataset}}} & \multicolumn{2}{c|}{$\Delta_\text{SP}$(\%) $\downarrow$}      & \multicolumn{2}{c}{$\Delta_\text{EO}$(\%) $\downarrow$}       \\ \cline{2-5} 
\multicolumn{1}{c}{}                         & Louvain             & Vanilla & Louvain             & Vanilla \\ \midrule
\textbf{Aminer-L}                                     & $\mathbf{5.26}_{\pm 0.47}$   & $5.95_{\pm 0.85}$   & $\mathbf{3.19}_{\pm 0.35}$   & $3.32_{\pm 0.60}$  \\
\textbf{Credit}                                     & $\mathbf{1.53}_{\pm 1.15}$   & $1.34_{\pm 0.63}$   & $\mathbf{0.88}_{\pm 0.63}$   & $1.67_{\pm 0.21}$  \\
\textbf{Pokec-n-G}                                     & $\mathbf{2.11}_{\pm 0.97}$   & $6.17_{\pm 1.07}$   & $\mathbf{5.55}_{\pm 3.08}$   & $14.20_{\pm 1.13}$  \\
\textbf{Pokec-n-R}                                     & $\mathbf{1.13}_{\pm 0.85}$   & $6.01_{\pm 0.42}$   & $\mathbf{3.66}_{\pm 1.76}$   & $5.96_{\pm 0.81}$  \\
\textbf{Pokec-z-G}                                     & $\mathbf{4.82}_{\pm 3.09}$   & $5.84_{\pm 0.69}$   & $\mathbf{3.84}_{\pm 0.91}$   & $4.99_{\pm 1.04}$  \\
\textbf{Pokec-z-R}                                     & $\mathbf{0.78}_{\pm 0.47}$   & $2.14_{\pm 0.69}$   & $\mathbf{0.62}_{\pm 0.41}$   & $5.44_{\pm 0.92}$  \\ \bottomrule
\end{tabular}
\caption{Fairness results for different graph partitioning strategies (Louvain vs Vanilla).}
\label{tab:partition_mini_louvain}
\vspace{-1em}
\end{table}

\begin{table}[H]
\centering
\tabcolsep=0.15cm
\footnotesize
\vspace{-1em}
\begin{tabular}{l|cc|cc}
\toprule
\multicolumn{1}{c}{\multirow{2}{*}{\textbf{Dataset}}} & \multicolumn{2}{c|}{$\Delta_\text{SP}$(\%) $\downarrow$}      & \multicolumn{2}{c}{$\Delta_\text{EO}$(\%) $\downarrow$}       \\ \cline{2-5} 
\multicolumn{1}{c}{}                         & Leiden             & Vanilla & Leiden             & Vanilla \\ \midrule
\textbf{Aminer-L}                                     & $\mathbf{5.53}_{\pm 0.75}$   & $5.95_{\pm 0.85}$   & $\mathbf{2.97}_{\pm 0.62}$   & $3.32_{\pm 0.60}$  \\
\textbf{Credit}                                     & $\mathbf{0.80}_{\pm 0.57}$   & $1.34_{\pm 0.63}$   & $\mathbf{0.49}_{\pm 0.39}$   & $1.67_{\pm 0.21}$  \\
\textbf{Pokec-n-G}                                     & $\mathbf{1.96}_{\pm 1.42}$   & $6.17_{\pm 1.07}$   & $\mathbf{4.48}_{\pm 1.37}$   & $14.2_{\pm 1.13}$  \\
\textbf{Pokec-n-R}                                     & $\mathbf{1.70}_{\pm 1.15}$   & $6.01_{\pm 0.42}$   & $\mathbf{2.90}_{\pm 2.13}$   & $5.96_{\pm 0.81}$  \\
\textbf{Pokec-z-G}                                     & $\mathbf{5.57}_{\pm 0.77}$   & $5.84_{\pm 0.69}$   & $\mathbf{4.01}_{\pm 1.25}$   & $4.99_{\pm 1.04}$  \\
\textbf{Pokec-z-R}                                     & $\mathbf{0.85}_{\pm 0.60}$   & $2.14_{\pm 0.69}$   & $\mathbf{1.10}_{\pm 0.76}$   & $5.44_{\pm 0.92}$  \\ \bottomrule
\end{tabular}
\caption{Fairness results for different graph partitioning strategies (Leiden vs Vanilla).}
\vspace{-1em}
\label{tab:partition_mini_leiden}
\end{table}

\begin{table}[H]
\centering
\tabcolsep=0.15cm
\footnotesize
\vspace{-1em}
\begin{tabular}{l|cc|cc}
\toprule
\multicolumn{1}{c}{\multirow{2}{*}{\textbf{Dataset}}} & \multicolumn{2}{c|}{$\Delta_\text{SP}$(\%) $\downarrow$}      & \multicolumn{2}{c}{$\Delta_\text{EO}$(\%) $\downarrow$}       \\ \cline{2-5} 
\multicolumn{1}{c}{}                         & Random             & Vanilla & Random             & Vanilla \\ \midrule
\textbf{Aminer-L}                                     & $\mathbf{4.98}_{\pm 0.24}$ & $5.95_{\pm 0.85}$ & $\mathbf{3.22}_{\pm 0.67}$ & $3.32_{\pm 0.60}$ \\
\textbf{Credit}                                       & $\mathbf{0.37}_{\pm 0.33}$ & $1.34_{\pm 0.63}$ & $\mathbf{0.31}_{\pm 0.23}$ & $1.67_{\pm 0.21}$ \\
\textbf{Pokec-n-G}                                    & $\mathbf{1.63}_{\pm 0.83}$ & $6.17_{\pm 1.07}$ & $\mathbf{3.73}_{\pm 1.00}$ & $14.2_{\pm 1.13}$ \\
\textbf{Pokec-n-R}                                    & $\mathbf{0.77}_{\pm 0.42}$ & $6.01_{\pm 0.42}$ & $\mathbf{1.84}_{\pm 1.68}$ & $5.96_{\pm 0.81}$ \\
\textbf{Pokec-z-G}                                    & $\mathbf{4.65}_{\pm 1.00}$ & $5.84_{\pm 0.69}$ & $\mathbf{2.39}_{\pm 0.63}$ & $4.99_{\pm 1.04}$ \\
\textbf{Pokec-z-R}                                    & $\mathbf{0.71}_{\pm 0.39}$ & $2.14_{\pm 0.69}$ & $\mathbf{0.75}_{\pm 0.61}$ & $5.44_{\pm 0.92}$ \\ \bottomrule
\end{tabular}
\caption{Fairness results for different graph partitioning strategies (Random vs Vanilla).}
\vspace{-1em}
\label{tab:partition_mini_random}
\end{table}

\section*{Appendix A.2}
The experimentals setting in empirical observations is the same as Appendix A.1.
The statistical information is based on the node features in each dataset.
We denotes nodes with degrees greater than 100 as higher-order nodes.

\section*{Appendix B}
We normalize the minority subgroups as $1$.
We denotes $\mathcal{V}_{h}$ as the higher-order node set.
For $\forall{v_j} \in \mathcal{V}_{h}$, if $|\mathbf{H}[j,s] = 1| > |\mathbf{H}[j,s] = 0|$:
\begin{equation}
    \begin{aligned}
        &\textbf{Higher-order Nodes}(\mathbf{H}[j,s] = 1) = \frac{|\mathbf{H}[j,s] = 1|}{|\mathbf{H}[j,s] = 0|}, \\
        &\textbf{Higher-order Nodes}(\mathbf{H}[j,s] = 0) = 1,
    \end{aligned}
\end{equation}
and, if $|\mathbf{H}[i,s] = 0| > |\mathbf{H}[j,s] = 1|$:
\begin{equation}
    \begin{aligned}
        &\textbf{Higher-order Nodes}(\mathbf{H}[j,s] = 1) = 1, \\
        &\textbf{Higher-order Nodes}(\mathbf{H}[j,s] = 0) = \frac{|\mathbf{H}[j,s] = 0|}{|\mathbf{H}[j,s] = 1|}.
    \end{aligned}
\end{equation}

For prediction results, if $\mathbb{P}(\hat{y} = 1, \mathbf{H}[i,s] = 1) > \mathbb{P}(\hat{y} = 1, \mathbf{H}[i,s] = 0)$:
\begin{equation}
    \begin{aligned}
        &\textbf{Prediction}(\hat{y} = 1,\mathbf{H}[i,s] = 1) = \frac{\mathbb{P}(\hat{y} = 1 | \mathbf{H}[i,s] = 1)}{\mathbb{P}(\hat{y} = 1 | \mathbf{H}[i,s] = 0)}, \\
        &\textbf{Prediction}(\hat{y} = 1,\mathbf{H}[i,s] = 0) = 1,
    \end{aligned}
\end{equation}
and if $\mathbb{P}(\hat{y} = 1, \mathbf{H}[i,s] = 0) > \mathbb{P}(\hat{y} = 1, \mathbf{H}[i,s] = 1)$:
\begin{equation}
    \begin{aligned}
        &\textbf{Prediction}(\hat{y} = 1,\mathbf{H}[i,s] = 1) = 1, \\
        &\textbf{Prediction}(\hat{y} = 1,\mathbf{H}[i,s] = 0) = \frac{\mathbb{P}(\hat{y} = 1 | \mathbf{H}[i,s] = 0)}{\mathbb{P}(\hat{y} = 1 | \mathbf{H}[i,s] = 1)}.
    \end{aligned}
\end{equation}

\section*{Appendix C}
\textbf{Lemma 1}
The similarity (measured by Euclidean Norm) between the distribution of original sensitive features and the distribution of sensitive features when they are mapping to node embeddings are lower than $\sqrt{n}$, whether using a vanilla GT or a GT with graph partitioning.
\begin{equation}
    \begin{aligned}
         &\Vert \mathbf{H}[:,s] - \hat{\textbf{A}}\mathbf{H}[:,s]\Vert_2 \leq \sqrt{n}, \\
         &\Vert \mathbf{H}[:,s] - \hat{\textbf{A}}'\mathbf{H}[:,s]\Vert_2 \leq  \sqrt{n}.
    \end{aligned}
\end{equation}

\textit{Proof.}
\begin{equation*}
    \begin{aligned}
        &\Vert \mathbf{H}[:,s] - \hat{\textbf{A}}\mathbf{H}[:,s]\Vert_2 \\
        = &\sqrt{\sum_{u \in \mathcal{V}} (\mathbf{H}[u, s] - \sum_{v \in \mathcal{V}} \hat{\mathbf{A}}[u,v] \mathbf{H}[v, s])^2} \\
        = &\sqrt{\sum_{u \in \mathcal{V}} \big( \sum_{v \in \mathcal{V}} \hat{\mathbf{A}}[u,v]( \mathbf{H}[u, s] - \mathbf{H}[v, s])\big)^2}. \\
        \leq &\sqrt{\sum_{u \in \mathcal{V}}(\sum_{v \in \mathcal{V}} \hat{\mathbf{A}}[u,v] 1)^2} \\
        = &\sqrt{\sum_{u \in \mathcal{V}} 1^2} \\
        = &\sqrt{n}.
    \end{aligned}
\end{equation*}

\begin{equation*}
    \begin{aligned}
        &\Vert \mathbf{H}[:,s] - \hat{\textbf{A}}'\mathbf{H}[:,s]\Vert_2 \\
        = &\sqrt{\sum_{u \in \mathcal{V}}\big( \sum_{v \in \mathcal{V}} \hat{\mathbf{A}}'[u,v]( \mathbf{H}[u, s] - \mathbf{H}[v, s])\big)^2}. \\
        \leq &\sqrt{n}.
    \end{aligned}
\end{equation*}
\rightline{$\square$}

\section*{Appendix D}
The statistics of these six datasets are shown in Table \ref{tab:datasets}.
\begin{table}[H]
  \centering
  \small
  \tabcolsep=0.1cm
  \begin{tabular}{ccccc}
  \toprule
  \textbf{Dataset}    & \textbf{\# Nodes} & \textbf{\# Edges} & \textbf{Sensitive feature} & \textbf{Label}\\
    \midrule
      \textbf{Credit}       & $30,000$  & $137,377$ & Age           & Credit \\
      \textbf{Pokec-z-R}    & $67,797$  & $882,765$ & Region        & Field \\
      \textbf{Pokec-z-G}    & $67,797$  & $882,765$ & Gender        & Field \\
      \textbf{Pokec-n-R}    & $66,569$  & $729,129$ & Region        & Field \\
      \textbf{Pokec-n-G}    & $66,569$  & $729,129$ & Gender        & Field \\
      \textbf{AMiner-L}     & $129,726$ & $591,039$ & Affiliation   & Field \\
    \bottomrule
  \end{tabular}
  \caption{Statistics of the experimental datasets. }
  \vspace{-1em}
  \label{tab:datasets}
\end{table}
\begin{itemize}
    \item \textbf{Credit} \cite{yeh2009the}: The edges of Credit dataset are based on similarities in their spending and payment patterns. 
    Age is considered the sensitive feature, while the label indicates whether an individual defaults on credit card payments in the following month.
    \item \textbf{Pokec-z} and \textbf{Pokec-n}~\cite{takac2012data}:Pokec-z and Pokec-n are constructed by sampling users from two provinces, Zilinsky and Nitriansky, respectively. 
    Each user is represented as a node with edges denoting friendship relations. 
    Node features are based on user profiles, including attributes such as living region, gender, spoken language, and age. 
    The sensitive features are living region and gender, with the classification task focusing on predicting users' working fields. 
    \item \textbf{AMiner-L}~\cite{wan2019aminer}: This dataset is a coauthor network constructed from the AMiner network. 
    The sensitive feature is the continent of each researcher's affiliation. 
    The associated task is to predict the primary research field of each researcher.
\end{itemize}

For datasets with more than two classes of ground truth labels for the node classification task, we simplify the problem by keeping the classes of labels $0$ and $1$ unchanged, while setting the classes of labels greater than $1$ to $1$. 
This simplification helps balance the classes and make the task more manageable.
Next, we partition the data by randomly selecting $25$\% of nodes as the validation set and another $25$\% as the test set, ensuring that each category of nodes remains balanced in these sets.
For the training set, we randomly select either $50$\% of nodes or $1,000$ nodes in each class of ground truth labels, depending on which is smaller. 
This partitioning strategy is consistent with the approach used in several prior studies~\cite{jiang2024chasing,luo2024fugnn,yang2024fairsin}.
By maintaining a balanced representation of classes in the training, validation, and test sets, we ensure that the model is evaluated fairly and consistently across different partitions. 
This strategy allows us to effectively measure the performance and fairness of the proposed FairGP method against the baseline approaches.

\section*{Appendix E}
Further comparison results for datasets \textbf{Pokec-n-R} and \textbf{Pokec-n-G} is shown in Table~\ref{tab:f_results}.
Notably, \textbf{Pokec-n-R} and \textbf{Pokec-n-G} are the same dataset examined under two different sensitive features.
The correlation results further validate FairGP's effectiveness in addressing the multi-sensitive feature challenge.

\begin{table}[H]
	\centering
    \footnotesize
	\caption{Comparison on utility (ACC and AUC) and fairness ( $\Delta_\text{SP}$ and $\Delta_\text{EO}$ ) in percentage (\%) in Pokec-n-R. $\uparrow$ denotes the larger, the better; $\downarrow$ denotes the opposite. The best results are red and bold-faced. The runner-ups are blue and underlined.}
    \vspace{-1em}
        \centering
        \tabcolsep=0.1cm
		\begin{tabular}{lcccc}
			\toprule
			\multirow{2}[2]{*}{\textbf{Methods}} & \multicolumn{4}{c}{\textbf{Pokec-n-R}}\\
            \cline{2-5}
			& ACC(\%) $\uparrow$ & AUC(\%) $\uparrow$  & $\Delta_\text{SP}$(\%) $\downarrow$   & $\Delta_\text{EO}$(\%) $\downarrow$  \\
			\midrule
			\textbf{GCN} & $66.53_{\pm 2.84}$ & $73.09_{\pm 0.28}$ & $6.57_{\pm 1.48}$ & $5.33_{\pm 0.42}$ \\
            \textbf{GAT} & $66.78_{\pm 1.30}$ & $71.00_{\pm 0.48}$ & $3.71_{\pm 0.15}$ & $7.50_{\pm 1.88}$ \\
            \textbf{APPNP} & $67.45_{\pm 1.18}$ & $69.80_{\pm 0.89}$ & $2.15_{\pm 0.23}$ & $4.35_{\pm 0.76}$ \\
			\textbf{FairGNN} & $65.29_{\pm 0.64}$ & $70.36_{\pm 2.06}$  & $5.30_{\pm 0.20}$ & $1.67_{\pm 0.17}$ \\
			\textbf{FairSIN} & \textcolor{red}{$\mathbf{69.34_{\pm 0.32}}$} & $71.57_{\pm 1.08}$ & \textcolor{blue}{\underline{$0.57_{\pm 0.19}$}} & \textcolor{blue}{\underline{$0.43_{\pm 0.41}$}} \\
            \textbf{FMP} & $67.36_{\pm 0.26}$ & $72.48_{\pm 0.26}$ & $0.66_{\pm 0.40}$ & $1.47_{\pm 0.87}$ \\
            \textbf{FUGNN} & $68.48_{\pm 0.07}$ & $72.83_{\pm 0.74}$ & $0.80_{\pm 0.31}$ & $1.03_{\pm 0.39}$ \\
            \textbf{DIFFormer} & $67.22_{\pm 1.55}$ & \textcolor{red}{$\mathbf{73.72_{\pm 0.50}}$} & $2.10_{\pm 0.10}$ & $2.33_{\pm 0.20}$ \\
            \textbf{SGFormer} & $64.55_{\pm 1.78}$ & $67.07_{\pm 1.54}$ & $4.14_{\pm 1.04}$  & $7.93_{\pm 0.66}$ \\
            \textbf{Polynormer} & $68.09_{\pm 0.23}$ & $71.43_{\pm 0.64}$  & $3.56_{\pm 1.45}$ & $2.17_{\pm 0.82}$ \\
            \textbf{CoBFormer} & $66.54_{\pm 0.64}$ & $70.79_{\pm 0.51}$ & $3.15_{\pm 1.01}$ & $5.40_{\pm 2.29}$ \\   
            \textbf{FairGT} & OOM & OOM & OOM & OOM \\
            \textbf{FairGP} & \textcolor{blue}{\underline{$68.86_{\pm 0.51}$}} & \textcolor{blue}{\underline{$73.35_{\pm 0.58}$}} & \textcolor{red}{$\mathbf{0.32_{\pm 0.12}}$} & \textcolor{red}{$\mathbf{0.20_{\pm 0.10}}$} \\
            \midrule
            \multirow{1}[1]{*}{} & \multicolumn{4}{c}{\textbf{Pokec-n-G}}\\
            \midrule
			\textbf{GCN} & $65.73_{\pm 0.51}$ & $70.60_{\pm 0.69}$ & $7.32_{\pm 1.35}$ & $12.03_{\pm 1.68}$ \\
            \textbf{GAT} & $66.37_{\pm 0.57}$ & $70.73_{\pm 0.25}$ & $11.21_{\pm 1.56}$ & $6.34_{\pm 2.00}$ \\
            \textbf{APPNP} & $66.91_{\pm 0.41}$ & $70.79_{\pm 0.52}$ & $8.83_{\pm 3.33}$ & $17.14_{\pm 3.97}$ \\
			\textbf{FairGNN} & $66.29_{\pm 1.37}$ & $70.43_{\pm 0.75}$ & $2.79_{\pm 0.75}$ & $2.91_{\pm 1.75}$ \\
			\textbf{FairSIN} & $65.91_{\pm 1.50}$ & $70.42_{\pm 1.69}$ & $2.64_{\pm 0.72}$ & \textcolor{blue}{\underline{$2.21_{\pm 0.32}$}} \\
            \textbf{FMP} & $65.33_{\pm 1.76}$ & $70.97_{\pm 0.62}$ & \textcolor{blue}{\underline{$2.32_{\pm 0.33}$}} & $2.31_{\pm 0.37}$ \\
            \textbf{FUGNN} & \textcolor{blue}{\underline{$66.95_{\pm 0.64}$}} & $70.77_{\pm 0.44}$ & $2.69_{\pm 0.12}$ & $2.80_{\pm 1.21}$ \\
            \textbf{DIFFormer} & \textcolor{red}{$\mathbf{67.00_{\pm 0.41}}$} & $70.96_{\pm 0.31}$ & $2.63_{\pm 0.40}$ & $4.31_{\pm 0.04}$ \\
            \textbf{SGFormer} & $65.78_{\pm 0.36}$ & \textcolor{red}{$\mathbf{71.29_{\pm 1.21}}$} & $2.87_{\pm 0.51}$ & $8.65_{\pm 1.49}$ \\
            \textbf{Polynormer} & $66.17_{\pm 0.59}$ & $70.55_{\pm 0.72}$ & $2.93_{\pm 1.44}$ & $9.48_{\pm 2.19}$ \\
            \textbf{CoBFormer} & $66.77_{\pm 0.86}$ & $70.71_{\pm 2.08}$ & $2.67_{\pm 1.64}$ & $2.47_{\pm 0.74}$\\   
            \textbf{FairGT} & OOM & OOM & OOM & OOM\\
            \textbf{FairGP} & $ 66.41_{\pm 0.41}$ & \textcolor{blue}{\underline{$71.24_{\pm 0.14}$}} & \textcolor{red}{$\mathbf{1.89_{\pm 0.12}}$} & \textcolor{red}{$\mathbf{1.64_{\pm 0.32}}$}\\
            \bottomrule
		\end{tabular}
\label{tab:f_results}
\end{table}

\section*{Appendix F}
Further ablation study for datasets \textbf{Credit}, \textbf{Pokec-n-R} and \textbf{Pokec-n-G} is shown in Table~\ref{tab:f_ablation}.
Even when compared to results using the mini-batch technique, FairGP shows a significant improvement in fairness, with enhancements of at least 50\% across these three datasets, and in some instances, exceeding 96\%.
These results validate our findings that graph partitioning improves the GT fairness. 

\begin{table}[H]
  \centering
  \scriptsize
  \tabcolsep=0.05cm
  \renewcommand{\arraystretch}{1.5}
  \begin{tabular}{lccccc}
    \toprule
    \textbf{Dataset} & \textbf{Metric} & \textbf{FairGP} & \textbf{w/o FM} & \textbf{w/o GP} & \textbf{w/o AO}\\
    \midrule
    \textbf{Credit} &  ACC $\uparrow$ & $\mathbf{78.13}_{\pm 0.08}$  & $76.40_{\pm 0.65}$ & $76.37_{\pm 0.45}$ & $77.64_{\pm 0.45}$ \\
    & AUC $\uparrow$ & $60.05_{\pm 2.98}$ & $59.15_{\pm 1.84}$ & $\mathbf{64.27}_{\pm 0.27}$ & $61.82_{\pm 1.40}$ \\
    & $\Delta_{\text{SP}}$ $\downarrow$ & $\mathbf{0.25}_{\pm 0.13}$ & $2.92_{\pm 0.95}$ & $1.34_{\pm 0.63}$ & $1.14_{\pm 0.08}$ \\
    & $\Delta_{\text{EO}}$ $\downarrow$ & $\mathbf{0.14}_{\pm 0.03}$ & $2.39_{\pm 0.35}$ & $1.67_{\pm 0.21}$ & $1.02_{\pm 0.49}$ \\
    \textbf{Pokec-n-R} & ACC $\uparrow$ & $\mathbf{68.86}_{\pm 0.51}$ & $66.55_{\pm 2.23}$ & $64.07_{\pm 0.09}$ & $68.50_{\pm 0.95}$ \\
    & AUC $\uparrow$ & $73.35_{\pm 0.58}$ & $70.79_{\pm 1.55}$ & $63.42_{\pm 2.01}$ & $\mathbf{73.95}_{\pm 1.70}$ \\
    & $\Delta_{\text{SP}}$ $\downarrow$ & $\mathbf{0.32}_{\pm 0.12}$ & $3.15_{\pm 1.00}$ & $6.01_{\pm 0.42}$ & $2.32_{\pm 0.19}$ \\
    & $\Delta_{\text{EO}}$ $\downarrow$ & $\mathbf{0.20}_{\pm 0.10}$ & $5.40_{\pm 2.28}$ & $5.96_{\pm 0.81}$ & $2.43_{\pm 1.68}$ \\
    \textbf{Pokec-n-G} & ACC $\uparrow$ & $66.41_{\pm 0.41}$ & $\mathbf{66.91}_{\pm 1.00}$ & $63.90_{\pm 0.13}$ & $65.59_{\pm 0.68}$ \\
    & AUC $\uparrow$ & $\mathbf{71.24}_{\pm 0.14}$ & $69.63_{\pm 1.68}$ & $63.54_{\pm 0.96}$ & $69.70_{\pm 0.51}$ \\
    & $\Delta_{\text{SP}}$ $\downarrow$ & $\mathbf{1.89}_{\pm 0.12}$ & $3.47_{\pm 0.32}$ & $6.17_{\pm 1.07}$ & $3.33_{\pm 0.15}$ \\
    & $\Delta_{\text{EO}}$ $\downarrow$ & $\mathbf{1.64}_{\pm 0.32}$ & $1.71_{\pm 0.24}$ & $14.20_{\pm 1.13}$ & $2.51_{\pm 1.15}$ \\
    \bottomrule
  \end{tabular}
  \caption{Ablation study of FairGP.}
  \label{tab:f_ablation}
  \vspace{-1em}
\end{table}

\section*{Appendix G}
\par The time required to compute the structure matrix $\mathbf{S}$ is short, and it can be reused across different hyperparameter experiments on the same dataset. 
In particular, FairGP selects only the principal eigenvectors using the Arnolodi Package, ensuring that remains consistently below. 
Table~\ref{tab:computing_s} shows the required runtime for computing $\mathbf{S}$.
\begin{table}[H]
  \centering
  \scriptsize
  \tabcolsep=0.25cm
  \renewcommand{\arraystretch}{1.5}
  \begin{tabular}{lccc}
    \toprule
    \textbf{Dataset} & \textbf{Pokec-z} & \textbf{Pokec-n} & \textbf{Aminer-L}\\
    \midrule
    \textbf{Computing S} & 8.43 & 7.91 & 10.95 \\
    \bottomrule
  \end{tabular}
  \caption{Runtimes of Computing S (s).}
  \label{tab:computing_s}
  \vspace{-1em}
\end{table}
\end{document}